\documentclass[conference]{IEEEtran}
\IEEEoverridecommandlockouts
\usepackage{cite}
\usepackage{amsmath,amssymb,amsfonts}
\usepackage{algorithmic}
\usepackage{graphicx}
\usepackage{textcomp}
\usepackage{xcolor}
\usepackage{multirow}
\usepackage{threeparttable}
\def\BibTeX{{\rm B\kern-.05em{\sc i\kern-.025em b}\kern-.08em
    T\kern-.1667em\lower.7ex\hbox{E}\kern-.125emX}}
\begin{document}

\title{On Reducing Activity with Distillation and Regularization for Energy Efficient Spiking Neural Networks}

\author{\IEEEauthorblockN{Thomas Louis}
\IEEEauthorblockA{IRT Saint Exupery, France\\
LEAT, Univ. Côte d’Azur, France\\
surname.lastname@univ-cotedazur.fr}
\and
\IEEEauthorblockN{Pierre-Emmanuel Novac}
\IEEEauthorblockA{LEAT, Univ. Côte d’Azur, France\\
surname.lastname@univ-cotedazur.fr}
\and
\IEEEauthorblockN{Benoit Miramond and Alain Pegatoquet}
\IEEEauthorblockA{LEAT, Univ. Côte d’Azur, France\\
surname.lastname@univ-cotedazur.fr}}

\makeatletter
\newcommand{\linebreakand}{%
  \end{@IEEEauthorhalign}
  \hfill\mbox{}\par
  \mbox{}\hfill\begin{@IEEEauthorhalign}
}
\makeatother

\author{
    \IEEEauthorblockN{Thomas Louis}
    \IEEEauthorblockA{IRT Saint Exupery, France\\
    LEAT, Univ. Côte d’Azur, France\\
    surname.lastname@univ-cotedazur.fr}
    \and
    \IEEEauthorblockN{Benoit Miramond}
    \IEEEauthorblockA{LEAT, Univ. Côte d’Azur, France\\
    surname.lastname@univ-cotedazur.fr}\\
    \linebreakand 
    \IEEEauthorblockN{Alain Pegatoquet}
    \IEEEauthorblockA{LEAT, Univ. Côte d’Azur, France\\
    surname.lastname@univ-cotedazur.fr}
    \and
    \IEEEauthorblockN{Adrien Girard}
    \IEEEauthorblockA{IRT Saint Exupery, France\\
    surname.lastname@irt-saintexupery.com}
}

\maketitle

\begin{abstract}
    Interest in spiking neural networks (SNNs) has been growing steadily, promising an energy-efficient alternative to formal neural networks (FNNs), commonly known as artificial neural networks (ANNs). Despite increasing interest, especially for Edge applications, these event-driven neural networks suffered from their difficulty to be trained compared to FNNs. To alleviate this problem, a number of innovative methods have been developed to provide performance more or less equivalent to that of FNNs. However, the spiking activity of a network during inference is usually not considered. While SNNs may usually have performance comparable to that of FNNs, it is often at the cost of an increase of the network's activity, thus limiting the benefit of using them as a more energy-efficient solution.

    In this paper, we propose to leverage Knowledge Distillation (KD) for SNNs training with surrogate gradient descent in order to optimize the trade-off between performance and spiking activity. Then, after understanding why KD led to an increase in sparsity, we also explored Activations regularization and proposed a novel method with Logits Regularization. These approaches, validated on several datasets, clearly show a reduction in network spiking activity ($-26.73\%$ on GSC and $-14.32\%$ on CIFAR-10) while preserving accuracy.

\end{abstract}


\section{Introduction}

Over the last few years, many compression methods such as quantization\cite{li_quantization_2022}, pruning\cite{liang_pruning_2021}, and KD \cite{gou_knowledge_2021} have been proposed for FNNs to reduce their energy consumption. In the meantime, SNNs have also been investigated\cite{yamazaki_spiking_2022} to reduce the energy consumption of machine learning algorithms.

Inspired by neuroscience, SNNs are event-driven and considered to be more energy-efficient than FNNs. Many studies have been proposed to train, compress\cite{sorbaro_optimizing_2020}, and deploy them \cite{davies_loihi_2018}\cite{admin_what_2022}\cite{abderrahmane_spleat_2022}\cite{painkras_spinnaker_2013}. However, most research efforts have been dedicated to quantization techniques, mainly for two reasons. First, deploying SNNs on specialized hardware, such as neuromorphic chips or ASICs/FPGAs, requires compression to accommodate their latency and memory footprint constraints. Secondly, the training of SNNs (which is not mandatory for quantization) is inherently more challenging compared to FNNs. Indeed, the non-differentiable spike function used by SNNs makes the training process more complex than for FNNs. However, since 2019, training SNN models has become possible using the surrogate gradient descent \cite{neftci_surrogate_2019}\cite{wu_direct_2019}, thus paving the way to compression methods that require training SNNs such as knowledge distillation (KD) \cite{xu_constructing_2023}. 

Using this direct training method, SNN-based algorithms having the same level of performance to FNNs have been proposed these last few years. Nevertheless, the energy consumed by SNNs is not always as low as expected. A crucial point, often neglected in recent works, is indeed the sparsity or the spiking activity of the network. The spiking activity is the proportion of spikes emitted by neurons during the inference. The lower the spiking activity (or the higher the sparsity), the better the energy efficiency \cite{lemaire_analytical_2023}. 

In this paper, we propose to leverage KD and regularization methods for SNNs to optimize the network sparsity while keeping the same level of performance. First, we investigate the benefits of Response-based KD methods to improve the SNNs sparsity. To further reduce the spiking activity, we then propose several regularization methods applied during the training phase. Our experiments have been performed with three different datasets: MNIST, GSC and CIFAR-10.

The main contributions of this paper are the following:
\begin{itemize}
    \item A novel "Logits Regularization" method that reduces the spiking activity while maintaining the accuracy.
    \item The first Sparsity impact analysis of three Response-based Distillation approaches for SNNs trained using surrogate gradient.
    \item In-depth comparison of sparsity and accuracy for the different regularization and KD methods, experiments being performed on three distinct datasets (MNIST, GSC and CIFAR-10).
\end{itemize}

The rest of this paper is organized as follows. Section \ref{RW} provides an examination of the existing knowledge landscape within the domain of spiking network's activity reduction. Then, Sections \ref{Methodology} and \ref{ES} describe the proposed methodology and our experimental setup, respectively. Subsequently, Section \ref{Results} presents and analyzes the obtained results. Finally, Section \ref{conclusion} encapsulates the synthesis of findings and potential avenues for future research.

\section{Related works}
\label{RW}

\subsection{Spiking Neural Networks}
SNNs employ neuron models to emulate the spiking behavior observed in the human brain's event-driven computation. Among the notable neuron models are the Integrate-and-Fire (IF) model and the Leaky Integrate-and-Fire (LIF) model. Both emulates the integration and firing mechanism of action potentials in the human brain. The IF model follows a straightforward concept, accumulating input currents until the current exceeds a threshold, which triggers a spike. The LIF model employs a leaky term to simulate the leakage of ions through the neuron's membrane, resulting in a gradual decay of the membrane potential over time.

%
%
%

Training SNNs poses distinctive challenges due to their event-driven dynamics and the non-differentiable nature of spiking behavior. Various strategies have been devised to tackle these hurdles and facilitate effective SNN training. A first approach is the FNN-to-SNN conversion. The FNN network is initially trained, then the acquired weights of the FNN are transferred to the SNN \cite{diehl_fast-classifying_2015}. Another prevalent method involves direct training through surrogate gradient descent \cite{neftci_surrogate_2019}, enabling the application of optimization algorithms based on back-propagation algorithm. In this paper, we only use the direct training method for our experiments.

\subsection{Knowledge Distillation}
\label{KD}

\begin{figure}[h]
    \centering
    \includegraphics[scale=0.2, trim={0 0 0 0cm}, clip]{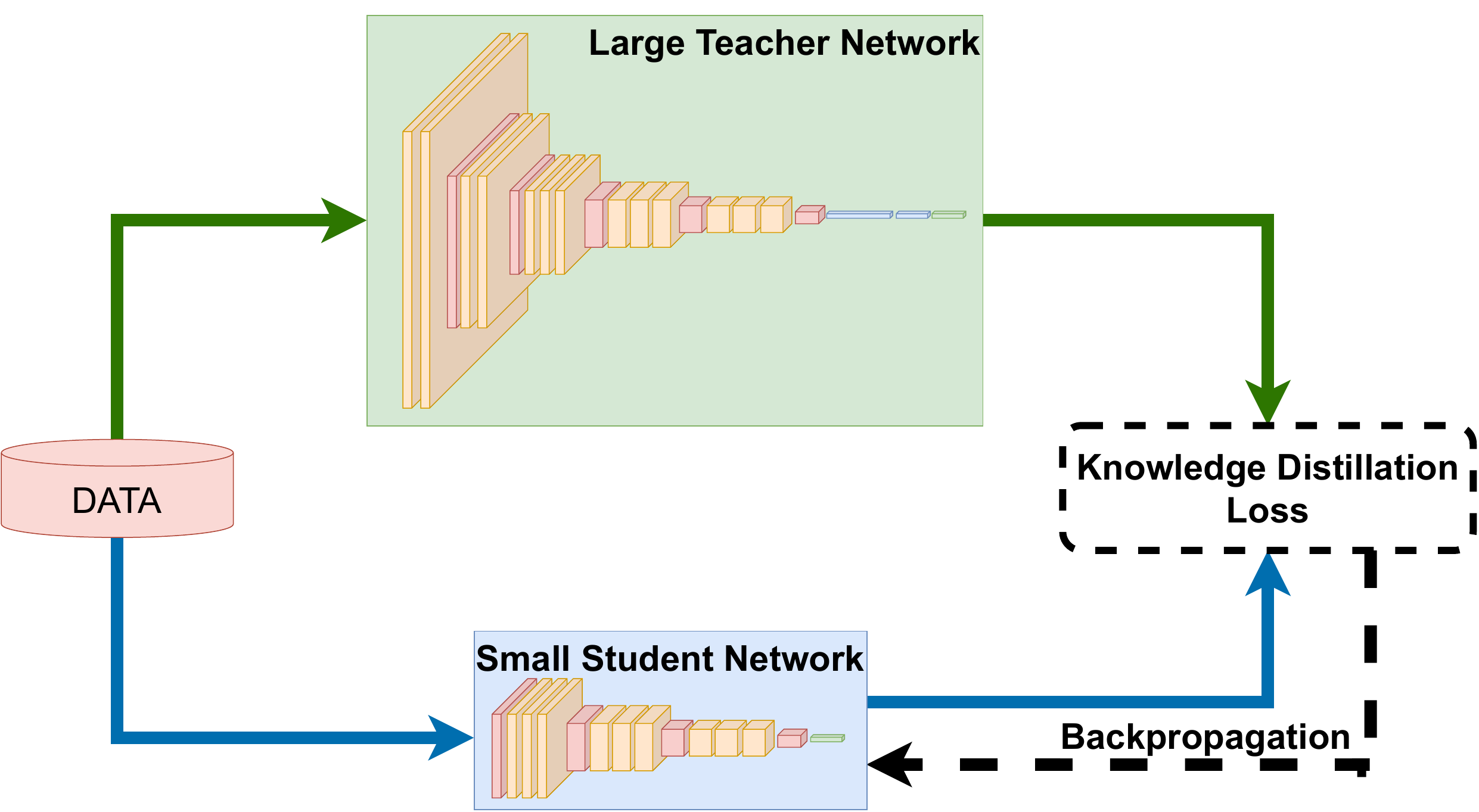}
    \caption{Knowledge Distillation explanation diagram}\label{fig:knowledge_distillation}
\end{figure}

KD \cite{gou_knowledge_2021} is a technique that trains a smaller neural network, called the student network, by leveraging the knowledge of a larger network, known as the teacher network. KD has been widely explored to reduce the size of FNNs while maintaining the performance. This area continues to garner attention, offering various techniques to distill knowledge effectively. One prominent method named Response-Based KD (shown in Figure \ref{fig:knowledge_distillation}) uses new labels generated by the teacher network, to compute a new loss and use it for back-propagation. The new total loss for Response-Based KD can be expressed as:
\begin{equation}
    \label{Ltotal}
    L_{\text{total}} = \alpha \times L_{\text{CLE}} + (1 - \alpha) \times L_{\text{KD}}
\end{equation}

With $L_{\text{CLE}}$ the Cross-Entropy Loss, $L_{\text{KD}}$ the distillation loss and $\alpha$ represents a weighting factor controlling the contribution of the distillation loss. 

To further enhance the distillation process, a temperature parameter is incorporated into the Softmax function. This temperature influences the sharpness of the output probability distribution of both student and teacher. Temperature therefore adds another layer of control to the distillation process, leading to improved results with optimized tuning. 

A more sophisticated KD approach, as presented in \cite{romero_fitnets_2015}, involves feature-based knowledge. In such a case, the student model learns  from feature maps of intermediate layers. This additional source of information enhances the transfer of knowledge, and is particularly useful with larger models.

While KD has been extensively studied and applied to FNNs, adapting it to SNNs is relatively recent in the literature. Early attempts, as in \cite{9412147}, involve incorporating the output spike train of the teacher SNN into the distillation process. In \cite{takuya_training_2021}, a model obtained through ANN-to-SNN conversion is fine-tuned using KD. In \cite{xu_constructing_2023} the authors proposed a novel method for constructing SNNs from ANNs using response-based and feature-based knowledge distillation methods, demonstrating the efficiency of the proposed KD SNN training method.

\subsection{Activations Regularization for SNNs}
In \cite{narduzzi_optimizing_2022}, the authors optimize the SNNs energy efficiency by applying Activations regularization to an FNN and then performing FNN-to-SNN conversion. Regularizers are based on the $\ell_{0}$, $\ell_{1}$, $\ell_{2}$ and $\ell_{p}$ norms as well as the Hoyer regularizer which is a ratio between $\ell_{1}$ and $\ell_{2}$ norms. Finally the Hoyer-Squares, which is a normalization of the $\ell_{0}$ norm, is also used.
The authors succeeded in regularizing two networks trained on MNIST and another one on CIFAR-10, then converting the FNN to SNN to achieve significant Spikerate reductions (see Table \ref{SOTA_table}).

In \cite{pellegrini_low-activity_2021}, a study was conducted to explore the effects of different modifications on the Spikerate of a three-layer convolutional SNN trained on the Google Speech Command dataset. The study examined four different variants: the use of non-dilated convolution, regularization of the Spikerate, switching from a LIF to an IF model, and freezing the leak of the LIF neurons. This study shows that it is possible to reduce the Spikerate of an SNN by applying various model transformations and different biological neurons or by applying regularization (see Table \ref{SOTA_table}). 

These studies collectively highlight the importance of energy optimization techniques in the context of SNNs, and shed light on various strategies to achieve this.

\section{Methodology}
\label{Methodology}

\subsection{Response-based Knowledge Distillation}
\label{KD_metho}
In this paper, we only focus on Response-based KD but we explore various techniques listed and explained below :

- Mean Squared Error (MSE) Knowledge Distillation:
this approach involves employing MSE in the KD loss function. The objective is to minimize the squared differences between the logits generated by the teacher and student networks. The MSE loss function is defined as follows:
\begin{equation}
    \label{LMSE}
    L_{\text{MSE}} = \| \text{logits}_{\text{t}}(x_i) - \text{logits}_{\text{s}}(x_i) \|^2
\end{equation}

Here, $\text{logits}_{\text{t}}$ and $\text{logits}_{\text{s}}$ represent the predicted logits (pre-softmax activations) of the teacher and student models, respectively, for the $x_i$ input.

- Soft Targets (ST) Knowledge Distillation:
the teacher's outputs are used as new labels (they are named Soft Targets or soft labels). In this method, a temperature $\tau$ is applied to the Softmax function of both the student and the teacher. The temperature parameter controls the smoothness of the probability distribution and therefore impacts the degree of knowledge transfered from the teacher to the student.
\begin{equation}
    \label{TSoftmax}
    p(x_i, \tau) = \frac{\exp(x_i / \tau)}{\sum_{j} \exp(x_j / \tau)}
\end{equation}
Where $p(x_i, \tau)$ is the probability distribution of the $x_i$ input and $\tau$ is the temperature parameter.
Accordingly, the loss term for ST KD is defined as follows:
\begin{equation}
    \label{LST}
    L_{\text{ST}} = \tau^2 * D_{\text{KL}}\left(p_t(x_i, \tau) \,||\, p_s(x_i, \tau)\right)
\end{equation}
Where $p_t$ and $p_s$ represent the probability distributions of the teacher and student models, respectively, for the $x_i$ input while $D_{\text{KL}}$ denotes the Kullback-Leibler divergence.

- Soft Targets Knowledge Distillation with Heterogeneous Temperature (ST-HET KD):
this technique proposes varying the temperature between the teacher and the student. The idea is to leverage the potential benefits of temperature heterogeneity in optimizing neural activity while preserving the performance. For ST-HET KD, the distillation loss is defined as:

\begin{equation}
    \label{LST-HET}
    L_{\text{ST-HET}} = \tau_t \times \tau_s \times D_{\text{KL}}\left(p_t(x_i; \tau_t) \,||\, p_s(x_i; \tau_s)\right)
\end{equation}
In addition to Eq. (\ref{LST}), $\tau_t$ and $\tau_s$ represent the temperature of the teacher and student models, respectively.

The distillation loss, whether MSE or ST, is added to the total loss function during training. The total loss shown in Eq. \ref{Ltotal} becomes a weighted combination of the standard loss and the distillation loss (Eq. $L_{\text{KD}}$ can be either Eq. \ref{LMSE}, or Eq. \ref{LST} or Eq. \ref{LST-HET}).

In this paper, we always set $\alpha$ to $0.1$ for Eq. \ref{LST} or Eq. \ref{LST-HET}, meaning that $90\%$ of the knowledge comes from the teacher's soften labels. There are several reasons to set $\alpha$ to $0.1$. First, this value is a common choice in the literature. Moreover, we did not observe significant improvements in our results when using a different value. Finally, for the sake of simplicity, we decided to keep the same $\alpha$ for all our experiments.

\subsection{Logits and Activations Regularization}
\label{Reg}

In this section, we focus on various regularization methods applied to SNNs models. These methods have been designed after an analysis of the results obtained with the ST-HET KD (see Eq. \ref{LST-HET}). This analysis shows that the Spikerate reduction come from a form of logits regularization (explained in Subsection \ref{Results_HET_KD}).

From now on, we process spikes as a tensor of binary activations and apply norms to it, thus giving a new term to add to the total loss. The goal of applying these regularization techniques is to encourage sparsity in the network, thereby improving the energy efficiency. In addition to Activations regularization, we propose a novel regularization method on the logits. To the best of our knowledge, this method has not been explored so far. Given that $t_i$ represents the activations of logits tensor and $n$ the number of activations or logits, the different types of regularization we applied are described below.

\begin{itemize}

    \item $\ell_1$ Activations Regularization: as binary activations (i.e., '0' or '1') are used in our case, the absolute value for Activations regularization is removed as it is unnecessary. Applying this norm leads to a model with less non-zero activations, thereby increasing sparsity.

          \begin{equation}
              \ell_{1norm} = \sum_{i=1}^{n} t_i
          \end{equation}
    
    \item $\ell_{2norm}^2$ Logits Regularization: to obtain a smoother optimization process, we use the squared $\ell_2$ norm, which is denoted by $\ell_2^2$ norm.
          \begin{equation}
              \ell_{2norm}^2 = \sum_{i=1}^{n} t_i^2
          \end{equation}

    \item $\ell_2$ Activations and Logits Regularization: Also known as Euclidean norm. It is worth noting that keeping the square is unnecessary in case of binary activations, but it has to be kept for logits.
            \begin{equation}
              \ell_{2norm} = \sqrt{\sum_{i=1}^{n} t_i^2}
          \end{equation}

\end{itemize}

In order to avoid being influenced by the number of layers, timesteps or neurons per tensor, a normalization is applied before adding the calculated norm(s), giving thus the following equation:

\begin{equation}
    \label{act_eq}
    \ell(\cdot) = \frac{1}{m} \sum_{j=1}^{m} \frac{\ell_{\text{norm}}(a_j)}{nT}
\end{equation}

Where $m$ is the number of layers in the model, $n$ is the number of neurons for the $j$-th layer, $T$ is the number of timesteps, $a_j$ is the activation tensor of the $j$-th layer, and $\ell_{\text{norm}}$ is the norm function that is applied.

Then, the loss function with the Activations regularization term can be written as:
\begin{equation}
    \label{reg_eq}
    L'(y,\hat{y}) = L(y,\hat{y}) + \lambda \ell(\cdot)
\end{equation}

where $L(y,\hat{y})$ is the original loss function, $\lambda$ is the regularization coefficient, and $\ell(\cdot)$ is one of the aforementioned regularization function (e.g. $\ell_1$ norm).

\section{Experimental Setup}
\label{ES}
\subsection{Datasets and models}
\label{data_and_mod}

In order to evaluate the benefits of the aforementioned methods, three distinct datasets have been used. The well-known MNIST dataset is used for image classification, containing 60,000 training images and 10,000 testing images. No data augmentation is applied to this dataset. The student SCNN model used with this dataset consists of two convolutional layers (16 and 64 filters of size 5x5), each layer being followed by 2x2 average pooling. The model ends with a fully connected layer of 10 neurons.

For an audio classification task, we use the Google Speech Command dataset (v0.02 without background noise) that features a diverse set of 35 spoken commands. We employed a combination of data augmentation techniques: Gaussian Noise ($\sigma$ at $1.75e-03$), Time Warping ($\sigma$ at $6.75e-02$) and Time Shifting ($\alpha$ at $1.0$). Gaussian Noise introduces random fluctuations to the data. Time Warping manipulates the timing of the input data, stretching or compressing it. Time Shifting randomly shiftes the timing of the data. Finaly, Mel-frequency cepstral coefficients (MFCC) are used to transform the raw audio data into a more compact representation. We used a sample rate of 16000 Hz and extracted 10 MFCC coefficients. The mel spectrogram was computed with 1024 FFT points and divided into 40 mel frequency bins. The analysis window length was set to 640 samples, with a hop length of 320 samples. The mel filter bank spanned from 20 Hz to 4000 Hz, and a padding of 320 samples was applied. The signal was not centered before computing the mel spectrogram.

The CIFAR-10 dataset has been used for large-scale visual recognition tasks. By comparing 60,000 RGB images (32×32 pixels) across 10 classes, CIFAR-10 represents a more complex task. For CIFAR-10, we used standard data augmentation techniques such as horizontal flipping and cropping (size=[32,32] and padding=[4,4]).

The results presented in subsequent sections are based on the average of three runs. The SNN student model (or baseline), the FNN teacher model, the spiking neuron type and the hyperparameters used for each experiment are shown in Table \ref{Models_prez}. 

\begin{table}[h]
    \centering
    \caption{Baseline Training Setup for each dataset}
    \begin{center}
        \begin{tabular}{|l|l|l|l|}
            \hline
            \textbf{Dataset} & \textbf{MNIST}                                           & \textbf{GSC}                                                                                                                                  & \textbf{CIFAR-10}                                                                                                                                                                  \\ \hline
            Baseline         & SCNN                    & SRes-8$_{45c}$ \cite{tang_deep_2018}                                                                                                          & SVGG-11                                                                                                                                                                            \\
            \hline
            Teacher          & VGG-13                                                   & ResNet-15                                                                                                                                     & VGG-16                                                                                                                                                                            \\
\hline
            Neuron/timesteps & IF/10                                                    & LIF/4                                                                                                                                         & LIF/4                                                                                                                                                                              \\
\hline
            epochs           & 50                                                       & 100                                                                                                                                           & 100                                                                                                                                                                                \\
\hline
            batch size       & 1000                                                     & 800                                                                                                                                           & 64                                                                                                                                                                                 \\
\hline
            Optimizer        & Adam                                                     & Adam                                                                                                                                          & SGD                                                                                                                                                                                \\

\hline
            Learning rate               &         1e-3                                                 & 2e-2                                                                                                                                          & 7.5e-2                                                                                                                                                              \\
            \hline
            Scheduler        & -                                  & \begin{tabular}[c]{@{}l@{}}Cosinus Annealing \\ T\_max = 100\end{tabular} & \begin{tabular}[c]{@{}l@{}}Step LR\\ gamma = 0.5\\ step size = 30\end{tabular} \\
\hline
            Data Augmentation                          & -                                                        & \begin{tabular}[c]{@{}l@{}}Gaussian Noise\\ Time Warping \\ Time Shifting\end{tabular}                                 & \begin{tabular}[c]{@{}l@{}}Horizontal Flip\\ Crop \end{tabular}                                                                            \\ \hline
            \end{tabular}
        \label{Models_prez}
    \end{center}
\end{table}

Models trained without any method to reduce the Spikerate are considered as our baseline models and are presented in Table \ref{baseline_table}. The number of parameters, the accuracy, the Spikerate, and the average total number of spikes for one sample are indicated for each student and teacher (KD only). It is noteworthy that these results serve as a reference point for our experiments. Our results will be compared to those of the baseline models (see Subsection \ref{Metrics}).

\begin{table}[h]
    \caption{Baselines Results for each model}
    \centering
        \begin{tabular}{lllll}
        \hline
        \textbf{Model}    & \textbf{Params} & \textbf{Accuracy}      & \textbf{Total Spikes} & \textbf{Spikerate}            \\ \hline
        \multicolumn{5}{c}{\textbf{MNIST}}                                                                              \\ \hline
        Teacher           & 28.3M           & 99,343\%          & \multicolumn{1}{c}{}  & \multicolumn{1}{c}{}          \\
        \textbf{Baseline} & \textbf{42k}    & \textbf{99,123\%} & \textbf{41201}        & \textbf{0.22}                \\ \hline
        \multicolumn{5}{c}{\textbf{GSC}}                                                                                \\ \hline
        Teacher           & 89k             & 94,66\%           & \multicolumn{1}{c}{}  & \multicolumn{1}{c}{\textbf{}} \\
        \textbf{Baseline} & \textbf{42k}    & \textbf{91,22\%}  & \textbf{24755}        & \textbf{0.44}                \\ \hline
        \multicolumn{5}{c}{\textbf{CIFAR-10}}                                                                           \\ \hline
        Teacher           & 15.3M           & 94,47\%           & \multicolumn{1}{c}{}  & \multicolumn{1}{c}{}          \\
        \textbf{Baseline} & \textbf{9.8M}   & \textbf{86,64\%}  & \textbf{24379}        & \textbf{0.10}                \\ \hline
        \end{tabular}
    \label{baseline_table}
\end{table}

\subsection{Metrics}\label{Metrics}
To comprehensively evaluate the efficiency of SNNs and the impact of each method, three metrics are used to capture various aspects of a network behavior and performance.

Accuracy Delta ($\delta_{\text{acc}_r}$):
$\delta_{\text{acc}_r}$ is defined as the relative error between the accuracy of the student model and the baseline model. It is calculated as follows:
\begin{equation}
    \label{acc_error}
    \text{$\delta_{\text{acc}_r}$} = \frac{\text{Acc}_{\text{model}} - \text{Acc}_{\text{baseline}}}{\text{Acc}_{\text{baseline}}}
\end{equation}
where $\text{Acc}_{\text{model}}$ the accuracy of the evaluated model and $\text{Acc}_{\text{baseline}}$ the accuracy of the baseline model.

Spike Count and Spikerate:
The total number of spikes generated by a SNN during inference is measured by the Spike Count metric. The Spikerate metric measures the average firing rate of neurons within a SNN. It is calculated as the ratio of the total Spike Count divided by the total number of IF/LIF neurons within the network:
\begin{equation}
    \label{Spikerate}
    \text{Sr} = \frac{\text{Sc}_{inf}}{\text{Tot}_{neurones}}
\end{equation}

With $\text{Sc}_inf$ the total number of spikes during one inference (i.e., on T timesteps) and ${\text{Tot}_{neurones}}$ the total number of IF/LIF neurons within the network. This means that the Spikerate is a value between [0,T], which is not always the case in the literature. The way the Spikerate is computated is indeed not always detailed. Sometimes, it represents the number of spikes per neurons per timestep. However, a lower Spikerate means decreasing neural activity and contributes to energy-efficient SNN implementations. In this paper, the Spikerate Relative Delta ($\delta_{\text{Sr}_r}$) which is defined as follows is used most of the time:
\begin{equation}
    \label{Spikerate_error}
    \text{$\delta_{\text{Sr}_r}$} = \frac{\text{Sr}_{\text{model}} - \text{Sr}_{\text{baseline}}}{\text{Sr}_{\text{baseline}}}
\end{equation}

where $\text{Sr}_{\text{model}}$ and $\text{Sr}_{\text{baseline}}$ being the Spikerate of the evaluated model and the baseline model, respectively.

This comprehensive set of metrics enables a thorough evaluation of the proposed regularization methods as well as their impact on the overall efficiency of SNNs.

\subsection{Qualia Framework}
In the context of this study, we use the Qualia framework \cite{noauthor_leat-edgequalia_2023} previously named MicroAI \cite{novac_quantization_2021}, a versatile tool designed for End-to-End training, quantization, and deployment of FNNs on embedded devices. Originally tailored for traditional FNNs, Qualia has recently extended its capabilities to accommodate SNNs with integration based on the SpikingJelly framework \cite{fang_spikingjelly_2023}.

To evaluate our approach, we have extended the Qualia framework by incorporating features dedicated to SNNs regularization and distillation. Furthermore, we have developed an additional feature enabling the calculation of the total number of spikes and the Spikerate for a spiking ResNet.

\section{Results}
\label{Results}

\subsection{Knowledge Distillation}
\label{Results_KD}

\subsubsection{MSE Knowledge Distillation Results}

The effect of MSE KD is evaluated for different distillation coefficients ($\alpha$) (see Eq. \ref{LMSE}). Obtained results are presented in terms of accuracy and Spikerate in Figure \ref{fig:MSE_KD}.

With MSE KD, an increase of $\alpha$ leads to different \text{$\delta_{\text{acc}_r}$} behaviours with a maximum gain for CIFAR-10 of $+0.86\%$ and a maximum $-2.63\%$ loss in $\delta_{\text{acc}_r}$ for GSC. However, for every dataset, $\delta_{\text{Sr}_r}$ starts with a positive value and then decreases to the baseline value. While MSE effectively aligns the logit distributions of student and teacher models, it fails to induce the student model to flatten its logits, thus failing to minimize its spikes count. These results provide a comparison of basic KD methods, but do not help reducing the SNN activity.

\begin{figure}[h]
    \centering
    \includegraphics[scale=0.42, trim={0cm 0cm 0cm 0cm}, clip]{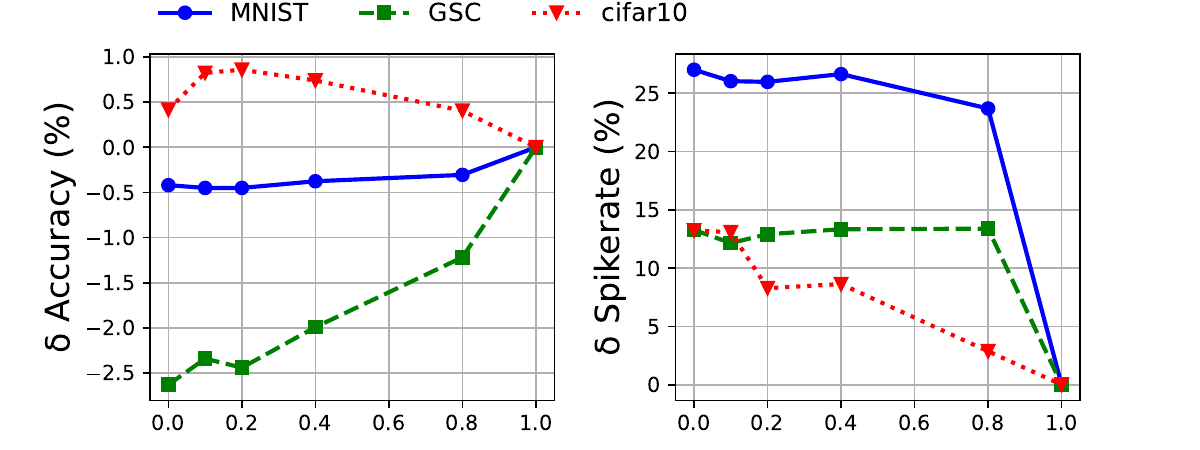}
    \caption{MSE Knowledge Distillation Results (Left: $\delta_{\text{acc}_r}$, Right: $\delta_{\text{Sr}_r}$)}
    \label{fig:MSE_KD}
\end{figure}

\subsubsection{Soft Targets Knowledge Distillation Results}
\label{Results_ST_KD}

As shown in Figure \ref{fig:ST_KD}, the ST KD method gives close \text{$\delta_{\text{acc}_r}$} results for all values of the temperature $\tau$. A trend can be observed for MNIST and GSC. As the temperature increases, the accuracy decreases. The opposite trend is observed with CIFAR-10, with a maximum gain of $+0.81\%$ at $\tau$ = 8. On the other hand, this method gives a higher $\delta_{\text{Sr}_r}$ than the student baseline, reaching after $\tau$ = 4 a Spikerate exceeding $+20\%$ for MNIST and GSC datasets. As it can be observed, the CIFAR-10 model seems to rapidly reach a plateau around $+5\%$ of gain in $\delta_{\text{Sr}_r}$. Again, there is no clear trend indicating that this method is efficient to reduce the Spikerate. As the smoothness of the student and the teacher are at the same level (i.e., the same temperature: see Eq. \ref{TSoftmax}), the distillation does not soften the student's logits, keeping large absolute values. To overcome this limitation, a more effective approach would be to encourage the student network to copy the flattened outputs of the teacher network, leading to a softening effect in the student's logits, which may potentially reduce the spiking activity.

\begin{figure}[h]
    \centering
    \includegraphics[scale=0.42, trim={0 0 0 0cm}, clip]{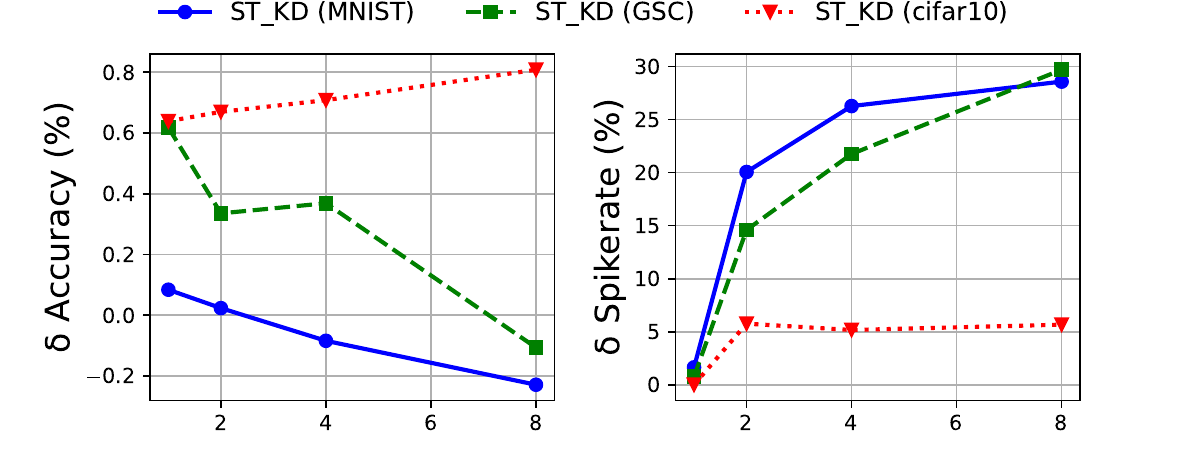}
    \caption{Soft Targets Knowledge Distillation Results (left : $\delta_{\text{acc}_r}$, right : $\delta_{\text{Sr}_r}$)}\label{fig:ST_KD}
\end{figure}

\subsubsection{Heterogeneous Temperature Knowledge Distillation Results}
\label{Results_HET_KD}

The results obtained with HET KD show a similar trend between datasets. In Figure \ref{fig:HET_KD}, it can be clearly observed that the Spikerate decreases as we approach [${\tau}_s$=1,${\tau}_t$=8]. The accuracy for all datasets tends to increase compared to the baseline, with the exception of CIFAR-10. As seen in (see Figure \ref{fig:HET_KD_cifar10}), the accuracy for CIFAR-10 begins to decrease at [${\tau}_s$=1,${\tau}_t$=4], reaching a maximum loss in \text{$\delta_{\text{acc}_r}$} of $-1.13\%$ at [${\tau}_s$=1,${\tau}_t$=8]. The \text{$\delta_{\text{acc}_r}$} divergence observed with CIFAR-10 shows a point where the teacher model produces too flat probabilities for the sharp student, leading to instability during training. For ${\tau}_s$ = 1, output probabilities of the student are generated from a normal Softmax function. For ${\tau}_t$ $>$ 4, the teacher produces probabilities that are too flat. The configuration ${\tau}_s$=1 and ${\tau}_t$ $>$ 4 leads the student to spread its probabilities, thus reducing logits range to smaller values. As shown in Figure \ref{fig:Distrib_MNIST_GSC_cifar10}, the last layers have the lowest Spikerates for HET KD. The closer you get to the beginning of the model, the less the Spikerate is reduced.

\begin{figure}[h]
    \centering
    \includegraphics[scale=0.40, trim={0cm 0cm 0cm 0cm}, clip]{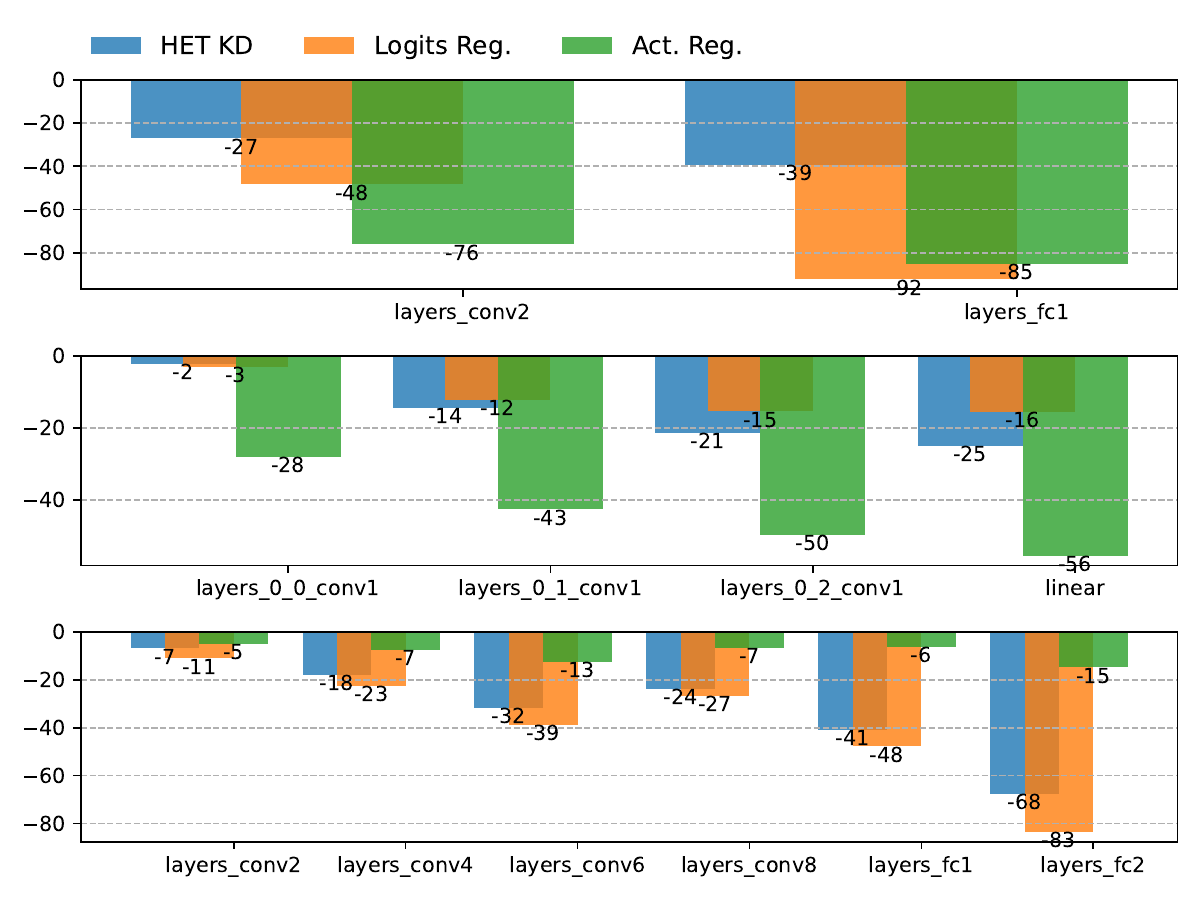}
    \caption{ST-HET KD, Logits Regularization and Activations Regularization $\delta_{\text{Sr}_r}$ (\%) per layer (Top: MNIST; Mid: GSC; Bot: CIFAR-10)}\label{fig:Distrib_MNIST_GSC_cifar10}
\end{figure}

\begin{figure}[h]
    \centering
    \includegraphics[scale=0.45, trim={1.5cm 1.5cm 1.5cm 2cm}, clip]{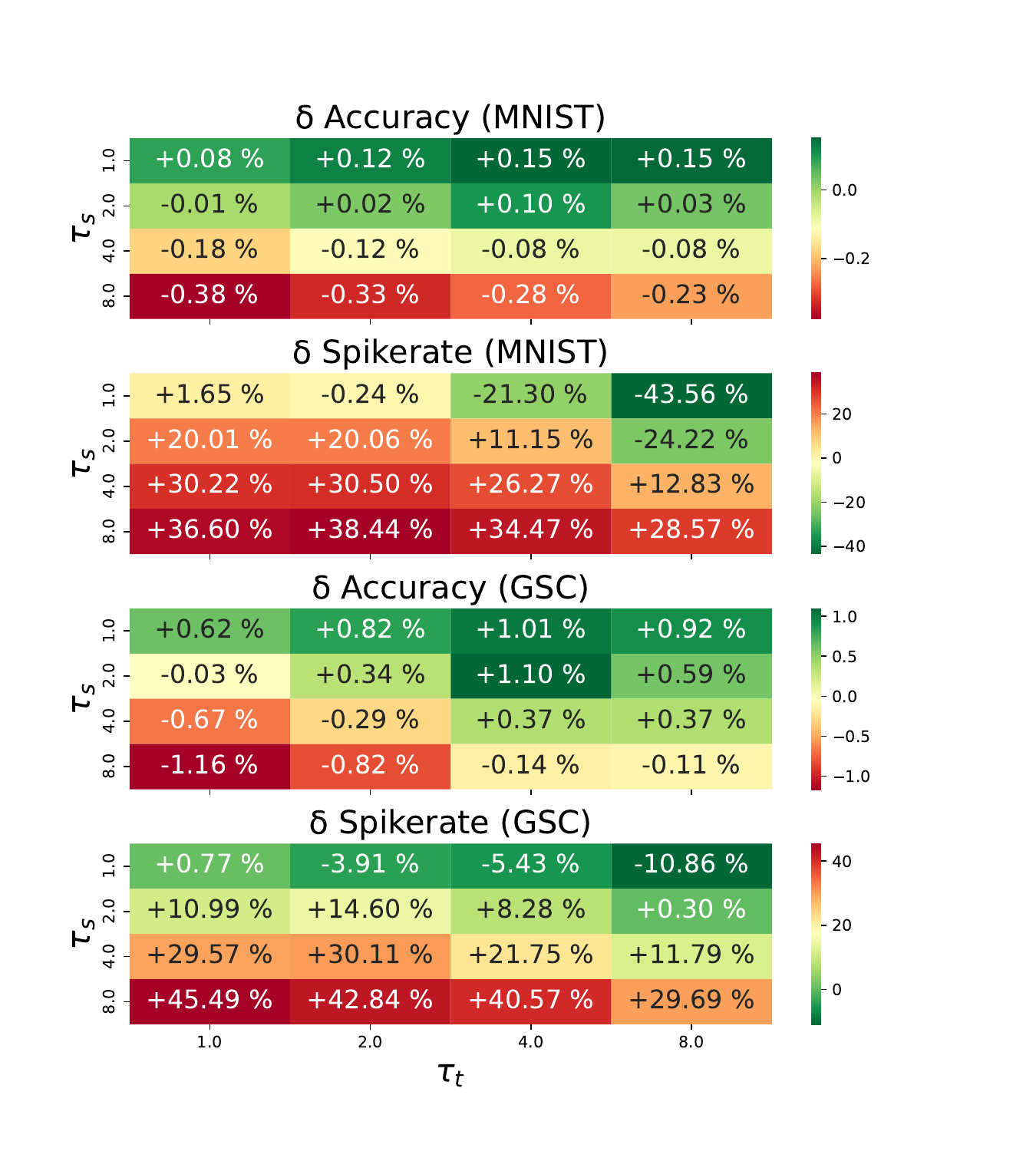}
    \caption{Heterogeneous Temperature Knowledge Distillation Results (Top: MNIST; Bot: GSC)}\label{fig:HET_KD}
\end{figure}

\begin{figure}[h]
    \centering
    \includegraphics[scale=0.35, trim={2cm 0cm 5cm 0cm}, clip]{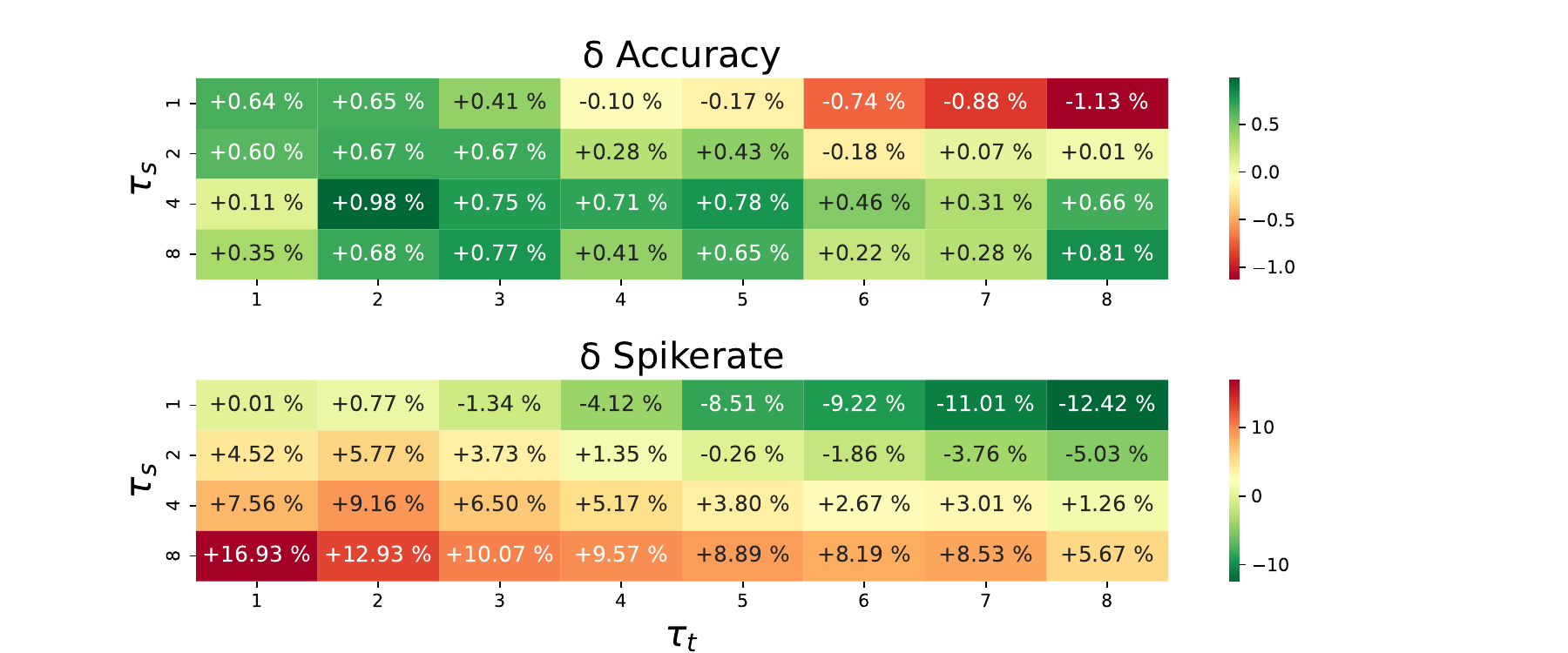}
    \caption{Heterogeneous Temperature Knowledge Distillation Results for CIFAR-10 (top: $\delta_{\text{acc}_r}$; bot: $\delta_{\text{Sr}_r}$)}\label{fig:HET_KD_cifar10}
\end{figure}

\subsection{Regularization Results}
\label{Results_reg}

\begin{figure*}[!h]
    \centering
    \includegraphics[scale=0.35, trim={0 0 0 0cm}, clip]{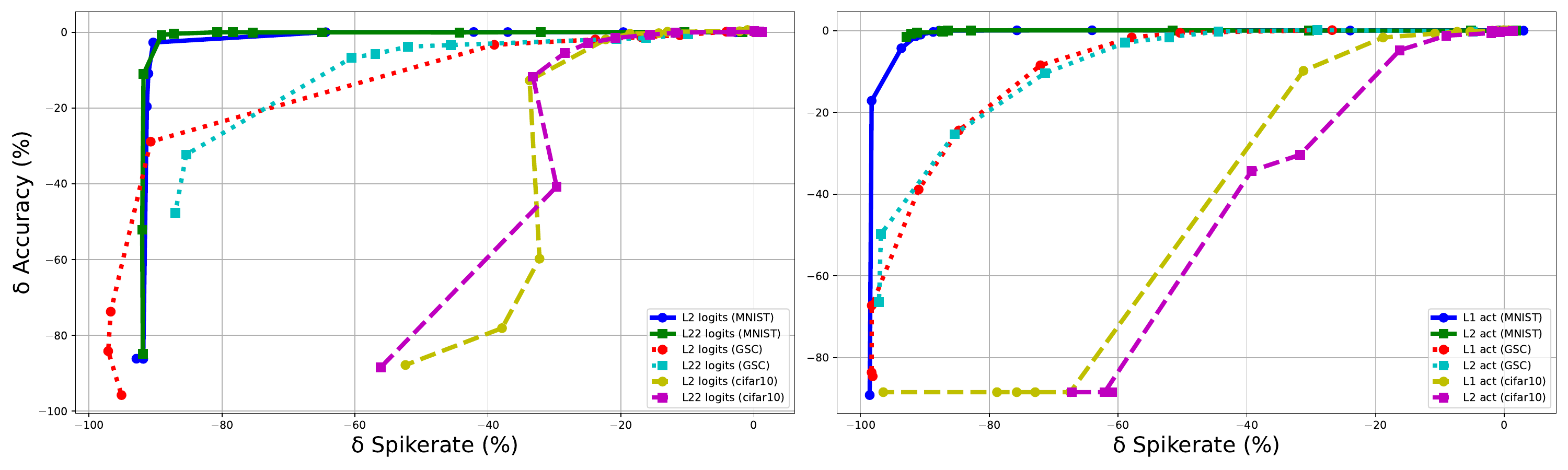}
    \caption{Logits (left) and Activations (right) Regularization Results for $\ell_{1norm}$, $\ell_{2norm}$ and $\ell_{2norm}^2$ norms}
    \label{fig:Logits_and_Act_MNIST_GSC_CIFAR-10}
\end{figure*}

In this section, different regularization methods are evaluated according to the model accuracy and sparsity. Comparisons are still conducted using $\delta_{\text{Acc}_r}$ and $\delta_{\text{Sr}_r}$. For each method, various regularization coefficients ($\lambda$) are explored. Figure \ref{fig:Logits_and_Act_MNIST_GSC_CIFAR-10} shows $\delta_{\text{Acc}_r}$ as a function of $\delta_{\text{Sr}_r}$ each point representing a value of $\lambda$. 
In these experiments, $\lambda$ belongs to [$1e-6$, $1e+6$] and [$1e-7$, $1e+2$] intervals for activations and logits regularization, respectively.

As it can be observed, $\ell_{2norm}$ and $\ell_{2norm}^2$ exhibit similar trends. For each dataset, the accuracy decreases when Logits regularization is used. Nevertheless, this accuracy drop starts at different $\delta_{\text{Sr}_r}$ values, around $-90\%$, $-40\%$ and $-20\%$ for MNIST, GSC and CIFAR-10, respectively. Moreover, the decrease of accuracy is quite more progressive for GSC and CIFAR-10 than for MNIST.

In the case of $\ell_{2norm}$ and $\ell_{1norm}$ Activations regularization, the accuracy starts to decrease from a higher $\delta_{\text{Sr}_r}$, around $-96\%$ and $-55\%$ for MNIST and GSC, respectively. Oddly, CIFAR-10 exhibits a decline from $-8\%$ in $\delta_{\text{Sr}_r}$ compared to the baseline. Logits regularization would therefore be more effective in this case. As shown in Figure \ref{fig:Distrib_MNIST_GSC_cifar10}, Logits regularization has higher impact than Activations regularization for each layer. We can also observe an expected behavior on the CIFAR-10 dataset: excessive regularization weight (i.e., large values of $\alpha$) leads to random accuracy and a Spikerate at $0\%$. This problem arises from the following reason. 
If the model regularizes too much, then the decrease of Spikerate can reach a critical point where there is not more spike emitted. In such a case, the gradient is zero and the model is obviously no longer able to learn.

Generally speaking, regularization on both logits and activations gives better results than KD. Moreover, Activations regularization seems to exhibit more reproducible and superior results. For Activations regularization, and as shown in Figure \ref{fig:Distrib_MNIST_GSC_cifar10}, it is clear that the Spikerate is reduced throughout the model, not only in the layers close to the end.

\subsection{Comparison with the State-of-the-Art}
\label{SOTA_comp}

In this section, we compare our methods with \cite{narduzzi_optimizing_2022}. This paper is the only one we have found in the literature that provides Spikerate metrics as well as accuracy results between a baseline model and models that have been subject to activity reduction techniques. Table \ref{SOTA_table} provides results from this work, along with ours. For our results, we selected the most effective configurations for KD, Logits Regularization (LogReg) and Activations Regularization (ActReg). 

\begin{table}[h]
\centering
\caption{SNNs sparsity optimization methods comparison}
\label{SOTA_table}
\begin{threeparttable}
    \begin{tabular}{llll}
    \hline
    \textbf{Work}                                       & \textbf{Arch./Params/T} & \textbf{Acc/$\delta_{\text{Acc}_r}$} & \textbf{*Tot. Spikes/$\delta_{\text{Sr}_r}$} \\ \hline
    \multicolumn{4}{c}{\textbf{MNIST}}                                                                                                                                  \\ \hline
    \text{\cite{narduzzi_optimizing_2022}}  & MLP/266k/100            & 97,55\%/-0,44\%                      & 29k/-90.93\%                               \\
    \text{\cite{narduzzi_optimizing_2022}}  & LeNet-5/?/100           & 97,92\%/-0,48\%                      & 411k/-89.72\%                                \\
    \textbf{**(KD)}                                  & CNN/42k/10              & 99.27\%/+0,15\%                      & 22k/-43,56\%                               \\
    \textbf{**(LogReg)}                              & CNN/42k/10              & 99.15\%/+0,03\%                      & 8k/-80,70\%                               \\
    \textbf{**(ActReg)}                              & CNN/42k/10              & 99.13\%/=                            & \textbf{5k/-87,81\%}                      \\ \hline
    \multicolumn{4}{c}{\textbf{GSC}}                                                                                                                                    \\ \hline
    \textbf{**(KD)}                                  & Res-8/41k/4             & \textbf{92,6\%/+0,92\%}              & 16,5k/-10,96\%                               \\
    \textbf{**(LogReg)}                              & Res-8/41k/4             & 90,71\%/-0,55\%                      & 16,6k/-9,88\%                                \\
    \textbf{**(ActReg)}                              & Res-8/41k/4             & 91,33\%/+0,12\%                      & \textbf{14k/-26,73\%}                      \\ \hline
    \multicolumn{4}{c}{\textbf{CIFAR-10}}                                                                                                                               \\ \hline
    \text{\cite{narduzzi_optimizing_2022}}  & LeNet-5/?/1000          & 51,76\%/+0,05\%                      & 14M/-76,48\%                                 \\
    \textbf{**(KD)}                                  & VGG11/10M/4            & \textbf{86,5\%/-0,17\%}              & 22k/-8,51\%                                \\
    \textbf{**(LogReg)}                              & VGG11/10M/4            & 86,47\%/-0,20\%                      & \textbf{21k/-14,32\%}                      \\
    \textbf{**(ActReg)}                              & VGG11/10M/4            & 86,41\%/-0,27\%                      & 23k/-7,3\%                                 \\ \hline
    \end{tabular}
    \begin{tablenotes}
        \footnotesize
        \item \textit{'*' Total number of Spikes. '**' means 'our work'. T for timesteps. '?' means unknown.}
      \end{tablenotes}
  \end{threeparttable}
    \end{table}

For MNIST, authors in \cite{narduzzi_optimizing_2022} leverage a Multilayer Perceptron (MLP) achieving a notable reduction of the total number of spikes ($-90.93\%$). Similarly, their LeNet-5 architecture reaches a decrease of the total number of spikes ($-89.72\%$). However, accuracy results for both models are quite low (below $98\%$). At iso-accuracy (not shown in this Table) we achieve very similar results ($-91.40\%$ at acc = $97.79\%$ with ActReg and $-89.09\%$ at acc = $98.41\%$ with LogReg). In contrast, our KD results show a slight increase in accuracy from the baseline, achieving a significant reduction in the total number of spikes ($-43.56\%$ and $-26.1\%$ respectively). Notably, our Logits and Activations regularization methods do not impact the accuracy performance, but significantly reduce the $\delta_{\text{Sr}_r}$ ($-80.70\%$ and $-87.81\%$, respectively).

For the GSC dataset, our KD approach with ResNet-8 achieves a slight gain in $\delta_{\text{Acc}_r}$ of $+0.92\%$, while reducing total spikes by $-10.96\%$. Comparatively, the LogReg does not give better results than KD, reaching only a $\delta_{\text{Sr}_r}$ reduction of $-9.88\%$. On the other hand, we can get competitive accuracies and a great reduction of $-26,73\%$ using ActReg.

Finally, for CIFAR-10, in \cite{narduzzi_optimizing_2022} the total number of Spikes is reduced from $61M$ to $14,4M$ while keeping an accuracy of $51.76\%$. Despite a nice reduction of the Spikerate, the accuracy remains very low. Moreover, there is a tremendous number of Spikes required for only one inference. Our KD approach with VGG11 achieves an accuracy of $86.5\%$ with a slight decrease from the baseline, coupled with a noteworthy reduction in total spikes ($-8.51\%$). ActReg approaches demonstrate a low reduction in accuracy from the baseline while achieving similar $\delta_{\text{Sr}_r}$ reduction ($-7.3\%$). Our LogReg methods gives the best results with a $\delta_{\text{Sr}_r}$ reduction of $-14.32\%$ while exhibiting a minimal reduction in accuracy.

These results highlight the efficiency of these methods across various datasets, demonstrating the potential for achieving competitive accuracy while significantly reducing the spiking activity. 

\section{Conclusion and Perspectives}
\label{conclusion}


In this study, we explored the benefits of regularization and KD techniques to reduce SNNs spiking activity. We have shown that distillation can reduce the SNNs Spikerate while maintaining the level of performance. Then, we explored regularization methods and we have shown that regularization of activations can significantly reduce the Spikerate of around $87\%$, $-26\%$ and $-7\%$ on MNIST, GSC and CIFAR-10 datasets, respectively. Finally, our new Logits regularization method can reduce the Spikerate ($-80\%$, $-10\%$ and $-14\%$ on MNIST, GSC and CIFAR-10 datasets, respectively) while maintaining reasonable accuracy.

As future works, and to further optimize the sparsity, we plan explore advanced combinations of regularization methods and KD. Specifically, applying feature-based KD techniques could offer insights into transferring more complex knowledge from FNNs to SNNs while applying regularization. The nature of this knowledge transfer raises questions, especially in the case of FNN to SNN distillation. These models have different architectures and learning mechanisms. SNNs use spiking activations, which are discrete signals, while FNNs use continuous activations. Therefore, it is unclear what form of knowledge is most transferable between the two types of models. We plan to explore different possibilities in order to determine what is the best knowledge to distill.

Seeking to accurately quantify the energy efficiency of our optimized SNN models, we intend to leverage the recent development of SPLEAT \cite{abderrahmane_spleat_2022}, a neuromorphic architecture for deploying SNN on ASIC and FPGA hardware. This will enable us to conduct real-world energy measurements and gain valuable insights into the operational energy consumption of our models.

%
%
%
%
%

\bibliographystyle{IEEEtran}
\bibliography{conference_101719}

\begin{thebibliography}{10}
\providecommand{\url}[1]{#1}
\csname url@samestyle\endcsname
\providecommand{\newblock}{\relax}
\providecommand{\bibinfo}[2]{#2}
\providecommand{\BIBentrySTDinterwordspacing}{\spaceskip=0pt\relax}
\providecommand{\BIBentryALTinterwordstretchfactor}{4}
\providecommand{\BIBentryALTinterwordspacing}{\spaceskip=\fontdimen2\font plus
\BIBentryALTinterwordstretchfactor\fontdimen3\font minus
  \fontdimen4\font\relax}
\providecommand{\BIBforeignlanguage}[2]{{%
\expandafter\ifx\csname l@#1\endcsname\relax
\typeout{** WARNING: IEEEtran.bst: No hyphenation pattern has been}%
\typeout{** loaded for the language `#1'. Using the pattern for}%
\typeout{** the default language instead.}%
\else
\language=\csname l@#1\endcsname
\fi
#2}}
\providecommand{\BIBdecl}{\relax}
\BIBdecl

\bibitem{li_quantization_2022}
\BIBentryALTinterwordspacing
C.~Li, L.~Ma, and S.~Furber, ``Quantization {Framework} for {Fast} {Spiking}
  {Neural} {Networks},'' \emph{Frontiers in Neuroscience}, vol.~16, 2022.
\BIBentrySTDinterwordspacing

\bibitem{liang_pruning_2021}
\BIBentryALTinterwordspacing
T.~Liang, J.~Glossner, L.~Wang, S.~Shi, and X.~Zhang,
  ``\BIBforeignlanguage{en}{Pruning and quantization for deep neural network
  acceleration: {A} survey},'' \emph{\BIBforeignlanguage{en}{Neurocomputing}},
  vol. 461, pp. 370--403, Oct. 2021.
\BIBentrySTDinterwordspacing

\bibitem{gou_knowledge_2021}
J.~Gou, B.~Yu, S.~J. Maybank, and D.~Tao, ``Knowledge {Distillation}: {A}
  {Survey},'' \emph{IJCV}, vol. 129, no.~6, pp. 1789--1819, Jun. 2021, arXiv:
  2006.05525.

\bibitem{yamazaki_spiking_2022}
\BIBentryALTinterwordspacing
K.~Yamazaki, V.-K. Vo-Ho, D.~Bulsara, and N.~Le,
  ``\BIBforeignlanguage{en}{Spiking {Neural} {Networks} and {Their}
  {Applications}: {A} {Review}},'' \emph{\BIBforeignlanguage{en}{Brain
  Sciences}}, vol.~12, no.~7, p. 863, Jul. 2022, number: 7 Publisher:
  Multidisciplinary Digital Publishing Institute.
\BIBentrySTDinterwordspacing

\bibitem{sorbaro_optimizing_2020}
\BIBentryALTinterwordspacing
M.~Sorbaro, Q.~Liu, M.~Bortone, and S.~Sheik, ``Optimizing the {Energy}
  {Consumption} of {Spiking} {Neural} {Networks} for {Neuromorphic}
  {Applications},'' \emph{Frontiers in Neuroscience}, vol.~14, 2020.
\BIBentrySTDinterwordspacing

\bibitem{davies_loihi_2018}
\BIBentryALTinterwordspacing
M.~Davies \emph{et~al.}, ``Loihi: {A} {Neuromorphic} {Manycore} {Processor}
  with {On}-{Chip} {Learning},'' \emph{IEEE Micro}, vol.~38, no.~1, pp. 82--99,
  Jan. 2018, conference Name: IEEE Micro.
\BIBentrySTDinterwordspacing

\bibitem{admin_what_2022}
\BIBentryALTinterwordspacing
admin, ``\BIBforeignlanguage{en-US}{What {Is} the {Akida} {Event} {Domain}
  {Neural} {Processor}?}'' May 2022. [Online]. Available:
  \url{https://brainchip.com/what-is-the-akida-event-domain-neural-processor-2/}
\BIBentrySTDinterwordspacing

\bibitem{abderrahmane_spleat_2022}
\BIBentryALTinterwordspacing
N.~Abderrahmane, B.~Miramond, E.~Kervennic, and A.~Girard, ``{SPLEAT}:
  {SPiking} {Low}-power {Event}-based {ArchiTecture} for in-orbit processing of
  satellite imagery,'' in \emph{2022 {IJCNN}}, Jul. 2022, pp. 1--10, iSSN:
  2161-4407.
\BIBentrySTDinterwordspacing

\bibitem{painkras_spinnaker_2013}
\BIBentryALTinterwordspacing
E.~Painkras \emph{et~al.}, ``{SpiNNaker}: {A} 1-{W} 18-{Core}
  {System}-on-{Chip} for {Massively}-{Parallel} {Neural} {Network}
  {Simulation},'' \emph{IEEE Journal of Solid-State Circuits}, vol.~48, no.~8,
  pp. 1943--1953, Aug. 2013, conference Name: IEEE Journal of Solid-State
  Circuits.
\BIBentrySTDinterwordspacing

\bibitem{neftci_surrogate_2019}
E.~O. Neftci, H.~Mostafa, and F.~Zenke, ``Surrogate {Gradient} {Learning} in
  {Spiking} {Neural} {Networks}: {Bringing} the {Power} of {Gradient}-{Based}
  {Optimization} to {Spiking} {Neural} {Networks},'' \emph{IEEE Signal
  Processing Magazine}, vol.~36, no.~6, pp. 51--63, 2019.

\bibitem{wu_direct_2019}
\BIBentryALTinterwordspacing
Y.~Wu, L.~Deng, G.~Li, J.~Zhu, Y.~Xie, and L.~Shi,
  ``\BIBforeignlanguage{en}{Direct {Training} for {Spiking} {Neural}
  {Networks}: {Faster}, {Larger}, {Better}},''
  \emph{\BIBforeignlanguage{en}{2019 AAAI}}, vol.~33, no.~01, pp. 1311--1318,
  Jul. 2019, number: 01.
\BIBentrySTDinterwordspacing

\bibitem{xu_constructing_2023}
\BIBentryALTinterwordspacing
Q.~Xu, Y.~Li, J.~Shen, J.~K. Liu, H.~Tang, and G.~Pan,
  ``\BIBforeignlanguage{en}{Constructing {Deep} {Spiking} {Neural} {Networks}
  {From} {Artificial} {Neural} {Networks} {With} {Knowledge} {Distillation}},''
  in \emph{\BIBforeignlanguage{en}{2023 {CVPR}}}, 2023, pp. 7886--7895.
\BIBentrySTDinterwordspacing

\bibitem{lemaire_analytical_2023}
E.~Lemaire, L.~Cordone, A.~Castagnetti, P.-E. Novac, J.~Courtois, and
  B.~Miramond, ``\BIBforeignlanguage{en}{An {Analytical} {Estimation}
  of {Spiking} {Neural} {Networks} {Energy} {Efficiency}},'' in
  \emph{\BIBforeignlanguage{en}{Neural {Information} {Processing}}}, ser.
  Lecture {Notes} in {Computer} {Science}, M.~Tanveer, S.~Agarwal, S.~Ozawa,
  A.~Ekbal, and A.~Jatowt, Eds.\hskip 1em plus 0.5em minus 0.4em\relax Cham:
  Springer International Publishing, 2023, pp. 574--587.

\bibitem{diehl_fast-classifying_2015}
\BIBentryALTinterwordspacing
P.~U. Diehl, D.~Neil, J.~Binas, M.~Cook, S.-C. Liu, and M.~Pfeiffer,
  ``Fast-classifying, high-accuracy spiking deep networks through weight and
  threshold balancing,'' in \emph{2015 {IJCNN}}, Jul. 2015, pp. 1--8, iSSN:
  2161-4407.
\BIBentrySTDinterwordspacing

\bibitem{romero_fitnets_2015}
A.~Romero, N.~Ballas, S.~E. Kahou, A.~Chassang, C.~Gatta, and Y.~Bengio,
  ``{FitNets}: {Hints} for {Thin} {Deep} {Nets},'' Mar. 2015, arXiv:1412.6550
  [cs].

\bibitem{9412147}
R.~K. Kushawaha, S.~Kumar, B.~Banerjee, and R.~Velmurugan, ``Distilling spikes:
  Knowledge distillation in spiking neural networks,'' in \emph{2020 ICPR},
  2021, pp. 4536--4543.

\bibitem{takuya_training_2021}
\BIBentryALTinterwordspacing
S.~Takuya, R.~Zhang, and Y.~Nakashima, ``\BIBforeignlanguage{en}{Training
  {Low}-{Latency} {Spiking} {Neural} {Network} through {Knowledge}
  {Distillation}},'' in \emph{\BIBforeignlanguage{en}{2021 {COOL}
  {CHIPS}}}.\hskip 1em plus 0.5em minus 0.4em\relax Tokyo, Japan: IEEE, Apr.
  2021, pp. 1--3.
\BIBentrySTDinterwordspacing

\bibitem{narduzzi_optimizing_2022}
\BIBentryALTinterwordspacing
S.~Narduzzi, S.~A. Bigdeli, S.-C. Liu, and L.~A. Dunbar, ``Optimizing {The}
  {Consumption} {Of} {Spiking} {Neural} {Networks} {With} {Activity}
  {Regularization},'' in \emph{2022 {ICASSP}}, May 2022, pp. 61--65, iSSN:
  2379-190X.
\BIBentrySTDinterwordspacing

\bibitem{pellegrini_low-activity_2021}
\BIBentryALTinterwordspacing
T.~Pellegrini, R.~Zimmer, and T.~Masquelier, ``Low-{Activity} {Supervised}
  {Convolutional} {Spiking} {Neural} {Networks} {Applied} to {Speech}
  {Commands} {Recognition},'' in \emph{2021 {IEEE} {SLT}}, Jan. 2021, pp.
  97--103.
\BIBentrySTDinterwordspacing

\bibitem{tang_deep_2018}
\BIBentryALTinterwordspacing
R.~Tang and J.~Lin, ``\BIBforeignlanguage{en}{Deep {Residual} {Learning} for
  {Small}-{Footprint} {Keyword} {Spotting}},'' in
  \emph{\BIBforeignlanguage{en}{2018 {IEEE} {ICASSP}}}.\hskip 1em plus 0.5em
  minus 0.4em\relax Calgary, AB: IEEE, Apr. 2018, pp. 5484--5488.
\BIBentrySTDinterwordspacing

\bibitem{noauthor_leat-edgequalia_2023}
\BIBentryALTinterwordspacing
``{LEAT}-{EDGE}/qualia.'' [Online]. Available:
  \url{https://github.com/LEAT-EDGE/qualia}
\BIBentrySTDinterwordspacing

\bibitem{novac_quantization_2021}
\BIBentryALTinterwordspacing
P.-E. Novac, G.~Boukli~Hacene, A.~Pegatoquet, B.~Miramond, and V.~Gripon,
  ``\BIBforeignlanguage{en}{Quantization and {Deployment} of {Deep} {Neural}
  {Networks} on {Microcontrollers}},'' \emph{\BIBforeignlanguage{en}{Sensors}},
  vol.~21, no.~9, p. 2984, Jan. 2021, number: 9 Publisher: Multidisciplinary
  Digital Publishing Institute.
\BIBentrySTDinterwordspacing

\bibitem{fang_spikingjelly_2023}
\BIBentryALTinterwordspacing
W.~Fang \emph{et~al.}, ``{SpikingJelly}: {An} open-source machine learning
  infrastructure platform for spike-based intelligence,'' \emph{Science
  Advances}, vol.~9, no.~40, p. eadi1480, Oct. 2023, publisher: American
  Association for the Advancement of Science.
\BIBentrySTDinterwordspacing

\end{thebibliography}

\vspace{12pt}
\color{red}
\end{document}